\documentclass[lettersize,journal]{IEEEtran}

\IEEEoverridecommandlockouts                              % This command is only needed if 
\usepackage{graphics} % for pdf, bitmapped graphics files
\usepackage{epsfig} % for postscript graphics files
\usepackage{mathptmx} % assumes new font selection scheme installed
\usepackage{times} % assumes new font selection scheme installed
\usepackage{amsmath} % assumes amsmath package installed
\usepackage{amssymb}  % assumes amsmath package installed
\usepackage{graphicx}
\usepackage{makecell}
\usepackage{caption}
\usepackage{mathrsfs}
\usepackage[ruled,vlined]{algorithm2e}
\usepackage[font={small}]{caption}
\usepackage[mathscr]{euscript}
\usepackage{bm}
\usepackage{pifont}% http://ctan.org/pkg/pifont
\usepackage{siunitx}
\usepackage{graphicx}
\usepackage{array}
\usepackage{float} 

\usepackage{array}
\newcommand{\PreserveBackslash}[1]{\let\temp=\\#1\let\\=\temp}

\newcolumntype{C}[1]{>{\PreserveBackslash\centering}p{#1}}
\newcolumntype{R}[1]{>{\PreserveBackslash\raggedleft}p{#1}}
\newcolumntype{L}[1]{>{\PreserveBackslash\raggedright}p{#1}}
\newcommand{\blockcomment}[1]{}

\usepackage[hidelinks]{hyperref}
\hypersetup{
    colorlinks=true,
    linkcolor=blue,
    urlcolor=black,
    citecolor=blue
}

\usepackage{multirow}
\usepackage{balance}
\usepackage{bbold}
\usepackage[utf8]{inputenc}
\usepackage[english]{babel}

\usepackage{etoolbox} % safe to include if not already loaded
\newcommand{\printbibliography}{%
\begin{thebibliography}{10}
\providecommand{\url}[1]{#1}
\csname url@samestyle\endcsname
\providecommand{\newblock}{\relax}
\providecommand{\bibinfo}[2]{#2}
\providecommand{\BIBentrySTDinterwordspacing}{\spaceskip=0pt\relax}
\providecommand{\BIBentryALTinterwordstretchfactor}{4}
\providecommand{\BIBentryALTinterwordspacing}{\spaceskip=\fontdimen2\font plus
\BIBentryALTinterwordstretchfactor\fontdimen3\font minus \fontdimen4\font\relax}
\providecommand{\BIBforeignlanguage}[2]{{%
\expandafter\ifx\csname l@#1\endcsname\relax
\typeout{** WARNING: IEEEtran.bst: No hyphenation pattern has been}%
\typeout{** loaded for the language `#1'. Using the pattern for}%
\typeout{** the default language instead.}%
\else
\language=\csname l@#1\endcsname
\fi
#2}}
\providecommand{\BIBdecl}{\relax}
\BIBdecl

\bibitem{gu_humanoid_2025}
\BIBentryALTinterwordspacing
Z.~Gu, J.~Li, W.~Shen, W.~Yu, Z.~Xie, S.~McCrory, X.~Cheng, A.~Shamsah, R.~Griffin, C.~K. Liu, A.~Kheddar, X.~B. Peng, Y.~Zhu, G.~Shi, Q.~Nguyen, G.~Cheng, H.~Gao, and Y.~Zhao., ``Humanoid locomotion and manipulation: Current progress and challenges in control, planning, and learning,'' \emph{IEEE/ASME Transactions on Mechatronics. DOI: 10.1109/TMECH.2025.3579247}, 2025.
\BIBentrySTDinterwordspacing

\bibitem{bretl_testing_2008}
\BIBentryALTinterwordspacing
T.~Bretl and S.~Lall, ``Testing static equilibrium for legged robots,'' vol.~24, no.~4, pp. 794--807.
\BIBentrySTDinterwordspacing

\bibitem{orsolino_feasible_2020}
\BIBentryALTinterwordspacing
R.~Orsolino, M.~Focchi, S.~Caron, G.~Raiola, V.~Barasuol, D.~G. Caldwell, and C.~Semini, ``Feasible region: An actuation-aware extension of the support region,'' vol.~36, no.~4, pp. 1239--1255.
\BIBentrySTDinterwordspacing

\bibitem{selvaggio2021autonomy}
M.~Selvaggio, M.~Cognetti, S.~Nikolaidis, S.~Ivaldi, and B.~Siciliano, ``Autonomy in physical human-robot interaction: A brief survey,'' \emph{IEEE Robotics and Automation Letters}, vol.~6, no.~4, pp. 7989--7996, 2021.

\bibitem{englsberger2011bipedal}
J.~Englsberger, C.~Ott, M.~A. Roa, A.~Albu-Sch{\"a}ffer, and G.~Hirzinger, ``Bipedal walking control based on capture point dynamics,'' in \emph{2011 IEEE/RSJ international conference on intelligent robots and systems}.\hskip 1em plus 0.5em minus 0.4em\relax IEEE, 2011, pp. 4420--4427.

\bibitem{dai_planning_2016}
\BIBentryALTinterwordspacing
H.~Dai and R.~Tedrake, ``Planning robust walking motion on uneven terrain via convex optimization,'' in \emph{2016 {IEEE}-{RAS} 16th International Conference on Humanoid Robots (Humanoids)}.\hskip 1em plus 0.5em minus 0.4em\relax {IEEE}, pp. 579--586.
\BIBentrySTDinterwordspacing

\bibitem{fernbach2020c}
P.~Fernbach, S.~Tonneau, O.~Stasse, J.~Carpentier, and M.~Taïx, ``C-croc: Continuous and convex resolution of centroidal dynamic trajectories for legged robots in multicontact scenarios,'' \emph{IEEE Transactions on Robotics}, vol.~36, no.~3, pp. 676--691, 2020.

\bibitem{wang2024online}
J.~Wang, S.~Kim, T.~S. Lembono, W.~Du, J.~Shim, S.~Samadi, K.~Wang, V.~Ivan, S.~Calinon, S.~Vijayakumar \emph{et~al.}, ``Online multicontact receding horizon planning via value function approximation,'' \emph{IEEE Transactions on Robotics}, vol.~40, pp. 2791--2810, 2024.

\bibitem{del_prete_fast_2016}
\BIBentryALTinterwordspacing
A.~Del~Prete, S.~Tonneau, and N.~Mansard, ``Fast algorithms to test robust static equilibrium for legged robots,'' in \emph{2016 {IEEE} International Conference on Robotics and Automation ({ICRA})}.\hskip 1em plus 0.5em minus 0.4em\relax {IEEE}, pp. 1601--1607.
\BIBentrySTDinterwordspacing

\bibitem{escande_planning_2013}
\BIBentryALTinterwordspacing
A.~Escande, A.~Kheddar, and S.~Miossec, ``Planning contact points for humanoid robots,'' vol.~61, no.~5, pp. 428--442.
\BIBentrySTDinterwordspacing

\bibitem{nozawa_three-dimensional_2016}
\BIBentryALTinterwordspacing
S.~Nozawa, M.~Kanazawa, Y.~Kakiuchi, K.~Okada, T.~Yoshiike, and M.~Inaba, ``Three-dimensional humanoid motion planning using {COM} feasible region and its application to ladder climbing tasks,'' in \emph{2016 {IEEE}-{RAS} 16th International Conference on Humanoid Robots (Humanoids)}.\hskip 1em plus 0.5em minus 0.4em\relax {IEEE}, pp. 49--56.
\BIBentrySTDinterwordspacing

\bibitem{audren_3-d_2018}
\BIBentryALTinterwordspacing
H.~Audren and A.~Kheddar, ``3-d robust stability polyhedron in multicontact,'' vol.~34, no.~2, pp. 388--403. [Online]. Available: \url{http://ieeexplore.ieee.org/document/8289420/}
\BIBentrySTDinterwordspacing

\bibitem{del_prete_zero_2018}
\BIBentryALTinterwordspacing
A.~Del~Prete, S.~Tonneau, and N.~Mansard, ``Zero step capturability for legged robots in multicontact,'' vol.~34, no.~4, pp. 1021--1034.
\BIBentrySTDinterwordspacing

\bibitem{ferrolho2023roloma}
H.~Ferrolho, V.~Ivan, W.~Merkt, I.~Havoutis, and S.~Vijayakumar, ``Roloma: Robust loco-manipulation for quadruped robots with arms,'' \emph{Autonomous Robots}, vol.~47, no.~8, pp. 1463--1481, 2023.

\bibitem{samadi_balance_2019}
S.~Samadi, S.~Caron, A.~Tanguy, and A.~Kheddar, ``Balance of humanoid robots in a mix of fixed and sliding multi-contact scenarios,'' in \emph{2020 IEEE International Conference on Robotics and Automation (ICRA)}, 2020, pp. 6590--6596.

\bibitem{roux_control_2021}
\BIBentryALTinterwordspacing
J.~Roux, S.~Samadi, E.~Kuroiwa, T.~Yoshiike, and A.~Kheddar, ``Control of humanoid in multiple fixed and moving unilateral contacts,'' in \emph{2021 20th International Conference on Advanced Robotics ({ICAR})}.\hskip 1em plus 0.5em minus 0.4em\relax {IEEE}, pp. 793--799.
\BIBentrySTDinterwordspacing

\bibitem{hirukawa_universal_2006}
\BIBentryALTinterwordspacing
H.~Hirukawa, S.~Hattori, K.~Harada, S.~Kajita, K.~Kaneko, F.~Kanehiro, K.~Fujiwara, and M.~Morisawa, ``A universal stability criterion of the foot contact of legged robots - adios {ZMP},'' in \emph{Proceedings 2006 {IEEE} International Conference on Robotics and Automation, 2006. {ICRA} 2006.}\hskip 1em plus 0.5em minus 0.4em\relax {IEEE}, pp. 1976--1983.
\BIBentrySTDinterwordspacing

\bibitem{caron_leveraging_2015}
\BIBentryALTinterwordspacing
S.~Caron, Q.~Cuong~Pham, and Y.~Nakamura, ``Leveraging cone double description for multi-contact stability of humanoids with applications to statics and dynamics,'' in \emph{Robotics: Science and Systems {XI}}.\hskip 1em plus 0.5em minus 0.4em\relax Robotics: Science and Systems Foundation.
\BIBentrySTDinterwordspacing

\bibitem{caron_zmp_2017}
\BIBentryALTinterwordspacing
S.~Caron, Q.-C. Pham, and Y.~Nakamura, ``{ZMP} support areas for multicontact mobility under frictional constraints,'' vol.~33, no.~1, pp. 67--80. [Online]. Available: \url{http://ieeexplore.ieee.org/document/7782743/}
\BIBentrySTDinterwordspacing

\bibitem{murooka_centroidal_2022}
\BIBentryALTinterwordspacing
M.~Murooka, M.~Morisawa, and F.~Kanehiro, ``Centroidal trajectory generation and stabilization based on preview control for humanoid multi-contact motion,'' vol.~7, no.~3, pp. 8225--8232. [Online]. Available: \url{https://ieeexplore.ieee.org/document/9807369/}
\BIBentrySTDinterwordspacing

\bibitem{rouxel_multicontact_2022}
\BIBentryALTinterwordspacing
Q.~Rouxel, K.~Yuan, R.~Wen, and Z.~Li, ``Multicontact motion retargeting using whole-body optimization of full kinematics and sequential force equilibrium,'' vol.~27, no.~5, pp. 4188--4198. [Online]. Available: \url{https://ieeexplore.ieee.org/document/9728754/}
\BIBentrySTDinterwordspacing

\bibitem{darvish_teleoperation_2023}
\BIBentryALTinterwordspacing
K.~Darvish, L.~Penco, J.~Ramos, R.~Cisneros, J.~Pratt, E.~Yoshida, S.~Ivaldi, and D.~Pucci, ``Teleoperation of humanoid robots: A survey,'' vol.~39, no.~3, pp. 1706--1727. [Online]. Available: \url{https://ieeexplore.ieee.org/document/10035484/}
\BIBentrySTDinterwordspacing

\bibitem{ramos2018dynamic}
J.~Ramos and S.~Kim, ``Humanoid dynamic synchronization through whole-body bilateral feedback teleoperation,'' \emph{IEEE Transactions on Robotics}, vol.~34, no.~4, pp. 953--965, 2018.

\bibitem{he2024learning}
T.~He, Z.~Luo, W.~Xiao, C.~Zhang, K.~Kitani, C.~Liu, and G.~Shi, ``Learning human-to-humanoid real-time whole-body teleoperation,'' in \emph{2024 IEEE/RSJ International Conference on Intelligent Robots and Systems (IROS)}.\hskip 1em plus 0.5em minus 0.4em\relax IEEE, 2024, pp. 8944--8951.

\bibitem{grandia_doc_2023}
\BIBentryALTinterwordspacing
R.~Grandia, F.~Farshidian, E.~Knoop, C.~Schumacher, M.~Hutter, and M.~Bächer, ``{DOC}: Differentiable optimal control for retargeting motions onto legged robots,'' vol.~42, no.~4, pp. 1--14.
\BIBentrySTDinterwordspacing

\bibitem{losey2018review}
D.~P. Losey, C.~G. McDonald, E.~Battaglia, and M.~K. O'Malley, ``A review of intent detection, arbitration, and communication aspects of shared control for physical human--robot interaction,'' \emph{Applied Mechanics Reviews}, vol.~70, no.~1, p. 010804, 2018.

\bibitem{freund2009postoptimal}
R.~M. Freund, ``Postoptimal analysis of a linear program under simultaneous changes in matrix coefficients,'' in \emph{Mathematical Programming Essays in Honor of George B. Dantzig Part I}.\hskip 1em plus 0.5em minus 0.4em\relax Springer, 2009, pp. 1--13.

\bibitem{dietrich2015overview}
A.~Dietrich, C.~Ott, and A.~Albu-Sch{\"a}ffer, ``An overview of null space projections for redundant, torque-controlled robots,'' \emph{The International Journal of Robotics Research}, vol.~34, no.~11, pp. 1385--1400, 2015.

\bibitem{chvatal1983linear}
V.~Chv{\'a}tal, \emph{Linear programming}.\hskip 1em plus 0.5em minus 0.4em\relax Macmillan, 1983.

\bibitem{bertrand_high-speed_2024}
\BIBentryALTinterwordspacing
S.~Bertrand, L.~Penco, D.~Anderson, D.~Calvert, V.~Roy, S.~{McCrory}, K.~Mohammed, S.~Sanchez, W.~Griffith, S.~Morfey, A.~Maslyczyk, A.~Mohan, C.~Castello, B.~Ma, K.~Suryavanshi, P.~Dills, J.~Pratt, V.~Ragusila, B.~Shrewsbury, and R.~Griffin, ``High-speed and impact resilient teleoperation of humanoid robots,'' in \emph{2024 {IEEE}-{RAS} 23rd International Conference on Humanoid Robots (Humanoids)}.\hskip 1em plus 0.5em minus 0.4em\relax {IEEE}, pp. 189--196.\BIBentrySTDinterwordspacing

\bibitem{orin_centroidal_2008}
\BIBentryALTinterwordspacing
D.~Orin and A.~Goswami, ``Centroidal momentum matrix of a humanoid robot: Structure and properties,'' in \emph{2008 {IEEE}/{RSJ} International Conference on Intelligent Robots and Systems}.\hskip 1em plus 0.5em minus 0.4em\relax {IEEE}, pp. 653--659.
\BIBentrySTDinterwordspacing

\bibitem{rakita2019shared}
D.~Rakita, B.~Mutlu, M.~Gleicher, and L.~M. Hiatt, ``Shared control--based bimanual robot manipulation,'' \emph{Science Robotics}, vol.~4, no.~30, p. eaaw0955, 2019.

\bibitem{corberes2025perceptive}
T.~Corbères, C.~Mastalli, W.~Merkt, J.~Shim, I.~Havoutis, M.~Fallon, N.~Mansard, T.~Flayols, S.~Vijayakumar, and S.~Tonneau, ``Perceptive locomotion through whole-body mpc and optimal region selection,'' \emph{IEEE Access}, vol.~13, pp. 69\,062--69\,080, 2025.

\end{thebibliography}

}

\usepackage{accents}
\usepackage[noend]{algpseudocode}

\usepackage{varwidth}
\newcommand\revText[1]{#1}

\newcommand\startRev{}
\newcommand\stopRev{}

\usepackage{mathtools}

\begin{document}

\title{\LARGE
% Posture Optimization for Robust Humanoid Multi-Contact Teleoperation
Stability-Aware Retargeting for Humanoid Multi-Contact Teleoperation
% Stability-Aware Teleoperation for Humanoid Multi-Contact
}

% \author{.$^{1,2}$ % <-this % stops a space
% \thanks{$^{1}$Author is with ------------------------------.}% <-this % stops a space
% \thanks{$^{2}$Author is with ------------------------------.}% <-this % stops a space
% \thanks{This work was supported through ---------------------------. Email: ------------} %
% }

\author{Stephen McCrory$^{1,2}$, Romeo Orsolino$^{3}$, Dhruv Thanki$^{1}$, Luigi Penco$^{1}$, Robert Griffin$^{1,2}$ % <-this % stops a space
\thanks{$^{1}$Author is with the Florida Institute for Human and Machine Cognition.}%
\thanks{$^{2}$Author is with the University of West Florida.}%
\thanks{$^{3}$Author is with Ocado Technology.}%
\thanks{This work was supported through ONR Grant No. N00014-19-1-2023. Email: \url{smccrory@ihmc.org}} %
}

\maketitle
\thispagestyle{empty}
\pagestyle{empty}

\begin{abstract}
Teleoperation is a powerful method to generate reference motions and enable humanoid robots to perform a broad range of tasks. However, teleoperation becomes challenging when using hand contacts and non-coplanar surfaces, often leading to motor torque saturation or loss of stability through slipping. We propose a centroidal stability-based retargeting method that dynamically adjusts contact points and posture during teleoperation to enhance stability in these difficult scenarios. Central to our approach is an efficient analytical calculation of the stability margin gradient. This gradient is used to identify scenarios for which stability is highly sensitive to teleoperation setpoints and inform the local adjustment of these setpoints. We validate the framework in simulation and hardware by teleoperating manipulation tasks on a humanoid, demonstrating increased stability margins. We also demonstrate empirically that higher stability margins correlate with improved impulse resilience and joint torque margin.
\end{abstract}

% \begin{IEEEkeywords}
% Whole-Body Motion Planning, Stability Analysis, Humanoid Robots, Virtual Reality Interfaces, Telerobotics and Teleoperation
% \end{IEEEkeywords}

\section{Introduction} %\IEEEPARstart{H}{umanoid} 
\IEEEPARstart{H}{umanoid} robots are rapidly becoming capable of coordinated, contact-rich interactions with their environments \cite{gu_humanoid_2025}. Teleoperation plays a crucial part in this development, enabling natural reference motions for a wide range of tasks to be quickly obtained from a human operator. However, performing multi-contact tasks, e.g. using hand contacts, remains unsupported by many frameworks. Addressing this requires the resolution of two competing objectives: track the operator, who provides desired motion setpoints, while maintaining balance. For many contact-rich motions the robot is on the boundary of actuation and friction limits, meaning stability should play a central role for the teleoperation of such scenarios.

\begin{figure}[t]
    \centering
    \includegraphics[trim={0cm 0 0cm 1.3cm},clip,width=0.95\columnwidth]{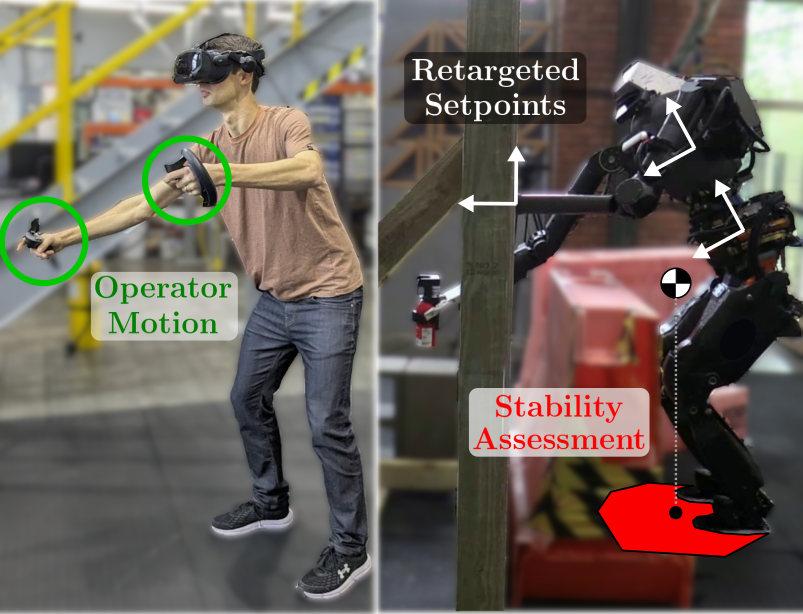}
    \caption{Robot performing a teleoperated manipulation task, in which the left hand contacts the wall to extend the reachable workspace of the robot and retrieve a canister. The robot's stability margin is improved by locally adjusting the posture and contact setpoints commanded from the operator.}
    \label{fig:title_figure}
\end{figure}

In this work, we propose a teleoperation scheme which retargets operator motions based on an actuation-aware 2d centroidal stability region \cite{bretl_testing_2008, orsolino_feasible_2020}. This retargeting acts both as a hard feasibility constraint and an objective to improve the stability margin, with the objective component being the primary focus of this work. In scenarios where the robot's stability margin changes with its posture or hand contact position (such as Fig. \ref{fig:title_figure}), this objective is ``activated'' and the operator's commanded motion is locally modified to improve stability. For cases when the stability margin is independent of posture, this adjustment is ``deactivated'' and the user has full control of the robot. In this way, our approach is a shared control scheme \cite{selvaggio2021autonomy} designed to balance the benefits of full user control and optimizing the robot's stability. % The retargeting is performed only on setpoints that do not affect the high-level task objectives, e.g. the grasping hand is not modified in Fig \ref{fig:title_figure}. 

Central to our approach is an efficient analytical calculation of the robot's centroidal stability margin gradient. The gradient is obtained by applying a Linear Program sensitivity property and is sufficiently fast for real-time application at kHz rates. Centroidal approaches have a rich history of enabling multi-contact locomotion \cite{englsberger2011bipedal, dai_planning_2016, fernbach2020c, wang2024online}. Online teleoperation is a particularly well-suited application for centroidal metrics as no trajectory preview is available, making it inherently a local motion planning problem. Additionally, the 2d stability region can visually provide the operator with an intuitive sense for the robot's balance. This is useful for contact conditions with multiple non-coplanar contacts in which the robot's stability would be otherwise difficult for an operator to intuit on-the-fly. Our experiments focus on such scenarios, including hand contacts in front (Fig.~\ref{fig:title_figure}), above (pushing downward) and behind (pushing backward) the robot.

% \blockcomment
% {
% Initial outline:
% - General statement about robustifying motions and avoiding falls. Also about a main advantage of humanoids over tracked/wheel/quadupred is using hand-holds to extend their workspace.
% - Briefly outline different stability metrics, ZMP, ICP, limit cycle
% - State what the paper is about. We are building on the work of xx and present yy.
% - Explain application to VR, etc.

% Other thoughts:
% - Discuss trade-off between stability and following the operator
% - In teleoperation, the future state isn't known, so any optimization must be local.
% - Advancing humanoid mobility necessarily involves pushing the robot to it's limits and using hand contacts where possible
% - Teleoperation has played an important role in this push, enabling robots to x, y, and z (cite Luigi's paper).
% - While retargeting
% }

\subsection{Related Work} 
One line of research in multi-contact stability analysis focuses on testing the feasibility of a given Center of Mass (CoM) state under unilateral and friction constraints. The seminal work of Bretl and Lall \cite{bretl_testing_2008} devised a method to efficiently compute the region of statically feasible CoM positions given a contact state. This 2d convex region, often referred to as the ``support region'' in contrast to the traditional ``support polygon,'' is valid on non-coplanar contact sets and is computed iteratively while maintaining an inner and outer estimate. Subsequent work in \cite{del_prete_fast_2016} reduced the dimensionality of the feasibility problem, resulting in faster solve times. This stability test (or similar variants) has been leveraged in planning non-gaited contact sequences over complex terrain \cite{escande_planning_2013, nozawa_three-dimensional_2016}. While these works enforce static stability through fixed contacts, others have aimed to relax this assumption. The stability region is extended to a 3d stability polytope in \cite{audren_3-d_2018}, which is robust to a set of user-specified CoM accelerations. Also through analysis of feasible CoM accelerations, in \cite{del_prete_zero_2018} a dynamic stability metric is formulated for a heuristic zero-step push recovery strategy. Robustness is explicitly optimized over a horizon in \cite{ferrolho2023roloma} through efficient calculation of the smallest unrejectable force. Modified support regions have also been formed in which a subset of support end-effectors are sliding or pushing an object through pre-defined trajectories \cite{samadi_balance_2019, roux_control_2021}.

A related line of research centers on the feasibility of contact wrenches given unilateral and friction constraints. Initial work in contact wrench feasibility analysis aimed towards extending the concept of Zero-Moment Point (ZMP) to non-coplanar contacts, pioneered by Hirukawa et al. \cite{hirukawa_universal_2006}. This concept was further developed through polytope projection methods to compute a set of feasible contact wrenches, the Contact Wrench Cone (CWC) \cite{caron_leveraging_2015}. The CWC has been used in a number of schemes to optimize centroidal trajectories, in which a robustness factor can readily be included \cite{dai_planning_2016, caron_zmp_2017}. In contrast to explicitly computing the feasible wrench space, other methods directly project a candidate contact wrench into the feasible space using a quadratic program \cite{murooka_centroidal_2022, rouxel_multicontact_2022}.

Stability modeling has largely not been leveraged in teleoperation, as many teleoperation systems are designed for relatively simple terrain or contact conditions \cite{darvish_teleoperation_2023}, with a number of notable exceptions. The robot's natural pendular frequency is leveraged to achieve bilateral feedback teleoperation in \cite{ramos2018dynamic}. Multi-contact teleoperation is performed in \cite{rouxel_multicontact_2022} through retargeting into the feasible dynamics of the robot. Whole-body teleoperation is demonstrated in \cite{he2024learning} by tracking learned setpoints through traditional whole-body control techniques. In \cite{grandia_doc_2023}, dynamic motions are transfered between robots of differing morphologies and realized through a whole-body Model-Predictive Control scheme. Lypanuov stability has been applied in shared controlled schemes to regulate the robot's exerted force \cite{losey2018review}.

\subsection{Contributions}
In this work, we apply state-of-the-art stability models to the problem of multi-contact teleoperation. Uniquely, our approach is designed to identify and improve the stability margin during teleoperation. Our contributions include:
\begin{itemize}
    \item An efficient analytical gradient of the centroidal stability region introduced in \cite{orsolino_feasible_2020}, obtained through Linear Program sensitivity analysis.
    \item A shared control teleoperation scheme which locally optimizes hand contact and posture to improve stability, when possible, or gives user full control otherwise.
    \item Simulation experiments for atypical contact conditions, demonstrating improved impulse resilience.
    \item Hardware demonstrations, including real-time controller and perception integration, to perform teleoperated manipulation, yielding increased joint torque margin.
\end{itemize}
\section{Background}
\subsection{Stability Region Linear Program}
We start by recapitulating the quasi-static CoM stability region for a robot with unilateral contacts \cite{bretl_testing_2008}. For a robot with CoM position $\mathbf{c}$, mass $m$, where contact point $k$ has position $\mathbf{p}_k$, force $\mathbf{f}_k$, friction cone $\mathscr{K}_k$, and gravity vector $\mathbf{g}$, the CoM is stable under the condition:
\begin{equation} \label{eq:StaticFeasibility}
\begin{aligned}
    \exists \, \mathbf{f} \;\; \textrm{s.t.}   \quad &  \mathbf{f}_k \in \mathscr{K}_k  \\
                                            \quad &   \sum_k \mathbf{f}_k = -m\mathbf{g} \\ 
                                            \quad &   \sum_k \mathbf{p}_k \times \mathbf{f}_k = -\mathbf{c} \times (m\mathbf{g}).
\end{aligned}
\end{equation}
Through a linearized friction cone constraint $\mathbf{Bf}\leq \mathbf{0}$ \cite{del_prete_fast_2016} and the observation that the stability test is independent of $\mathbf{c}_z$, boundary values of the region are computed through a Linear Program (LP). Additionally, an actuation constraint can be included which locally models the joint torque bounds \cite{orsolino_feasible_2020}
\begin{equation}\label{eq:ActuationConstraint}
    \mathbf{g}(\mathbf{q}) - \boldsymbol{\tau}^{+} \leq \mathbf{J}(\mathbf{q})^T_c\mathbf{f} \leq \mathbf{g}(\mathbf{q}) - \boldsymbol{\tau}^{-}.
\end{equation}
This 2d region enables an efficient model of the robot's stability at the current contact state. A stability margin $m$, which is the minimum distance from the CoM to the boundary of the region, models the robot's impulse resilience at the current contact state \cite{orsolino_feasible_2020}. Vertex $i$ on this actuation-aware stability region is solved by querying along a direction $\mathbf{a}_i \in \mathbb{R}^2$, representing the extreme feasible CoM position $\mathbf{c}^\ast_{xy}$ achievable in that direction.
\begin{equation} \label{eq:CoMLP}
\begin{aligned}
    &\max_{\mathbf{x}} \;\; a_i = \mathbf{a}_i^T \mathbf{c}_{xy} \\
    & \hspace{1.45cm}  \textrm{such that:} \\
    &\overbrace
    {
    \begin{bmatrix}
        (\mathbf{I}_3\ldots\mathbf{I}_3) & \mathbf{0} & \mathbf{0} & \mathbf{0} \\
        ([\mathbf{p}_0]_\times\ldots) & -m[\mathbf{g}]_\times & \mathbf{0} & \mathbf{0} \\
        \mathbf{J}(\mathbf{q})^T_c & \mathbf{0} & -\mathbf{I} & \mathbf{0} \\
        \mathbf{J}(\mathbf{q})^T_c & \mathbf{0}  & \mathbf{0} & \mathbf{I}
    \end{bmatrix}
    }
    ^{\mathbf{A}(\mathbf{p},\mathbf{q})} 
    \overbrace{\begin{bmatrix}
        \mathbf{f} \\ 
        \mathbf{c} \\ 
        \mathbf{m_\tau^+} \\
        \mathbf{m_\tau^-}
    \end{bmatrix}
    }
    ^{\mathbf{x}} 
    \,\, = \,\,
    \overbrace{
    \begin{bmatrix}
        -m\mathbf{g} \\
        \mathbf{0} \\
        \mathbf{g}(\mathbf{q}) - \bm{\tau}^+ \\
        \mathbf{g}(\mathbf{q}) - \bm{\tau}^-
    \end{bmatrix}
    }
    ^{\mathbf{b}(\mathbf{q})} \\[6pt]
 & \hspace{1.45cm} \, \mathbf{Bf}\leq\mathbf{0}, \ \mathbf{m}^+_\tau, \mathbf{m}^-_\tau\geq\mathbf{0} \\
\end{aligned}
\end{equation}

\begin{figure}
    \centering
    \includegraphics[trim={0cm 0.5cm 0cm 0cm},clip,width=0.7\columnwidth]{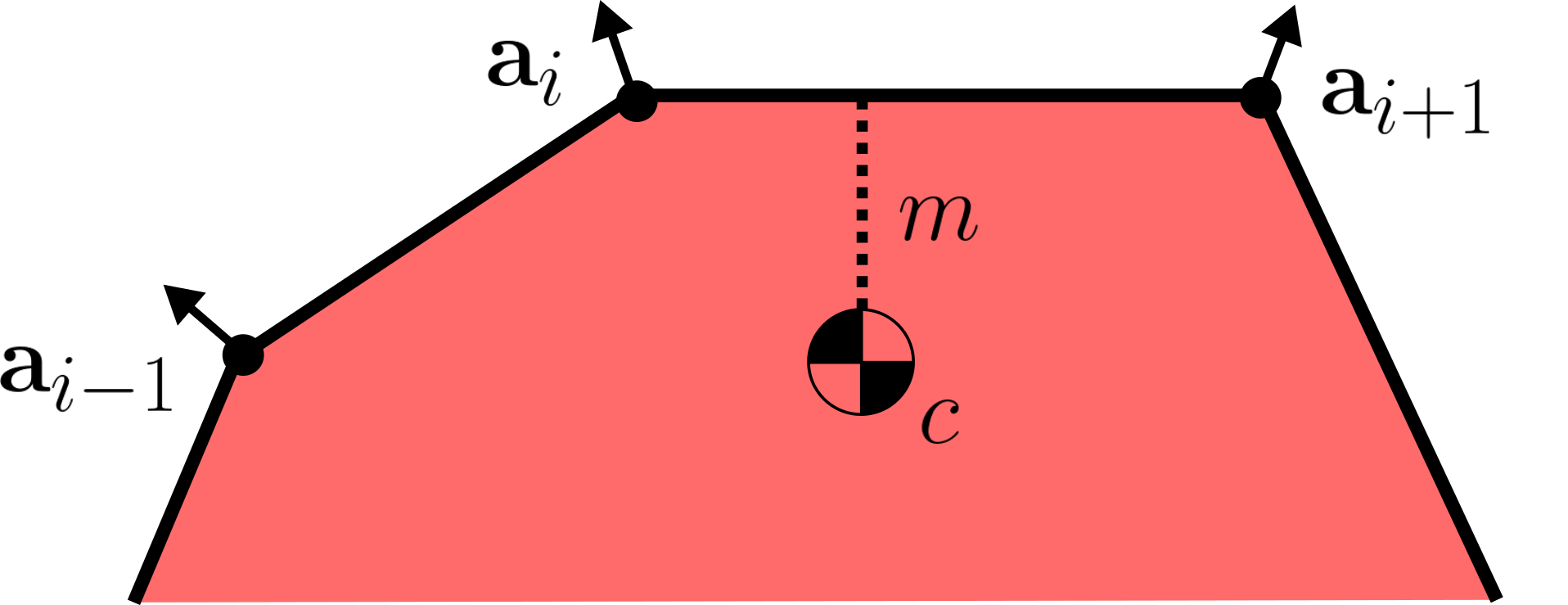}
    \caption{Depiction of the stability region. Vertex $i$ corresponds to a direction $\mathbf{a}_i$ and is computed by Eq.~(\ref{eq:CoMLP}). The stability margin $m$ is the minimum distance from the CoM $\mathbf{c}$ to the region boundary.}
    \label{fig:OverheadStability}
\end{figure}

Where the robot has $n_q$ total degrees of freedom, $n_j$ actuated joints, $n_c$ contact points, $\mathbf{q} \in \mathbb{R}^{n_q}$ is the whole-body configuration vector, $\mathbf{g} \in \mathbb{R}^{n_j}$ is the vector of static gravity-induced torques, $\mathbf{J}_c \in \mathbb{R}^{n_j \times 3n_c}$ is a stacked matrix of contact Jacobians, and $\boldsymbol{\tau}^{+/-} \in \mathbb{R}^{n_j}$ are the upper/lower joint torque bounds. The operator $[\cdot]_\times$ designates a skew symmetric matrix such that $[\mathbf{p]_\times\,f}=\mathbf{p}\times\mathbf{f}$. Finally, the slack variables $\mathbf{m}^{\pm}_\tau$ are the joint torque margins corresponding to the computed contact force $\mathbf{f}^\ast$. Fig.~\ref{fig:OverheadStability} depicts an overhead view of an example stability region, in which vertex $i$ represents the vertex corresponding to query direction $\mathbf{a}_i$.
\subsection{Linear Program Sensitivity Analysis}
\label{sec:SensitivityBackground}
Here we review a general property of Linear Programs stated in \cite{freund2009postoptimal}. A sensitivity analysis can be performed for a standard form Linear Program where the constraint matrix $\mathbf{A}$ is linearly parameterized by a scalar $\theta$ and matrices $\mathbf{F},\mathbf{G}$.
\begin{equation}\label{eq:ParametericLP}
\begin{aligned}
    \max_\mathbf{x} \quad & a(\theta) = \mathbf{a}^T \mathbf{x}  \\
    \textrm{s.t.} \quad & \overbrace{(\mathbf{F} + \mathbf{G}\theta)}^{\mathbf{A}(\theta)}\mathbf{x} = \mathbf{b} \\
    \quad  & \mathbf{x} \geq \mathbf{0} \\
\end{aligned}
\end{equation}
For a Linear Program with this structure, the derivative of $a^\ast(\theta)$, the optimal solution of Eq.~(\ref{eq:ParametericLP}), is given by
\begin{equation}\label{eq:SensitivityFormula}
    \frac{\partial a^\ast}{\partial \theta} \biggr\rvert_{\theta=\theta_0} = - \mathbf{y}^{\ast T} \mathbf{G} \mathbf{x}^{\ast},
\end{equation}
where $\mathbf{x}^{\ast}$, $\mathbf{y}^{\ast}$ are the primal and dual optimal solutions to Eq.~(\ref{eq:ParametericLP}) at $\theta=\theta_0$, respectively.
\section{Stability Analysis}
In this section we compute the gradient of $a^*_i$, the optimal solution to Eq.~(\ref{eq:CoMLP}), with respect to two quantities: contact point position $\mathbf{p}_k$ and whole-body configuration $\mathbf{q}$. The quantities $\nabla a^\ast_i(\mathbf{p}_k)$ and $\nabla a^\ast_i(\mathbf{q})$ represent the optimal contact point and whole-body configuration motions which respectively move the stability region vertex $i$ towards the direction $\mathbf{a}_i$. We begin by applying Eq.~(\ref{eq:SensitivityFormula}) to compute the sensitivity of a stability region vertex along the direction $\mathbf{a}_i$ with respect to these quantities. To match the parameterized form Eq.~(\ref{eq:ParametericLP}), a time-linearization of $\mathbf{A}$ is used:
\begin{equation}\label{eq:ConstraintMatrixLinearization}
    \mathbf{A}(t)\big|_{t=t_0} \approx \mathbf{A}(t_0) + \frac{\partial \mathbf{A}}{\partial t} \Delta t.
\end{equation}
Note additional slack variables are included so Eq.~(\ref{eq:CoMLP}) is standard form, but are omitted here for brevity. We assume the constraint vector is approximately constant, $\dot{\mathbf{b}}(\mathbf{q})\approx 0$. Physically this means that the gravitational torques do not change significantly with time compared to the contact Jacobians, which we confirm empirically (see Appendix). From Eq.~(\ref{eq:SensitivityFormula}), this yields the scalar velocity of vertex $i$ along direction $\mathbf{a}_i$ as
\begin{equation}\label{eq:VertexSensitivity}
    \frac{\partial a^\ast_i}{\partial t} \biggr\rvert_{t=t_0} = - \mathbf{y}^{\ast T} \frac{\partial \mathbf{A}}{\partial t}\mathbf{x}^{\ast},
\end{equation} \startRev
where the time-derivative of $\mathbf{A}$ is given by 
\begin{equation}\label{eq:ConstraintDerivative}
    \frac{\partial \mathbf{A}}{\partial t} = 
    \begin{bmatrix}
        \;\mathbf{0} & \mathbf{0} & & \\[-4pt]
        \;(\ldots [\mathbf{\dot{p}}_k]_\times \ldots) & & \ddots & \\[-1pt]
        \;\dot{\mathbf{J}}^T_c(\mathbf{v}) & & \ddots & \\[2pt]
        \;\mathbf{\dot{J}}^T_c(\mathbf{v}) & & & \mathbf{0} \\
    \end{bmatrix}.
\end{equation}
To leverage Eq.~(\ref{eq:VertexSensitivity}) in the calculation of $\nabla a^\ast_i(\mathbf{z})$, where $\mathbf{z}$ is a placeholder for $\mathbf{p}_k$ or $\mathbf{q}$, we consider the general property
\begin{equation}\label{eq:GradientTotalDerivativeRelationship}
    \frac{da^\ast_i}{dt} = \mathbf{\dot{z}} \cdot \nabla a^\ast_i(\mathbf{z}).
\end{equation}
The $j^{th}$ entry of $\nabla a^\ast_i(\mathbf{z})$ can be computed by evaluating Eq.~(\ref{eq:GradientTotalDerivativeRelationship}) for $\mathbf{\dot{z}}=\hat{\dot{\mathbf{z}}}_j$, where $\hat{\dot{\mathbf{z}}}_j$ is the $j^{\text{th}}$ standard basis vector,
\begin{equation}\label{eq:GenericGradienRelatedToTimeDerivative}
    \nabla a^\ast_i (\mathbf{z}) = 
    \begin{bmatrix}
        \frac{\partial a^\ast_i}{\partial t}(\dot{\mathbf{z}} = \hat{\dot{\mathbf{z}}}_0) \;\; \ldots \;\; \frac{\partial a^\ast_i}{\partial t}(\dot{\mathbf{z}} = \hat{\dot{\mathbf{z}}}_{n}) \\
    \end{bmatrix}^T.
\end{equation}
Therefore $\nabla a_i^\ast$ can be computed through a series of evaluations of Eq.~(\ref{eq:VertexSensitivity}), with $\mathbf{z}=\mathbf{p}_k$ or $\mathbf{z}=\mathbf{q}$ to compute the gradient with respect to the $k^{\text{th}}$ contact position or whole-body configuration, respectively. Although the robot's contact positions are a function of the whole-body configuration, we chose to consider the gradients separately since the contact position gradient is significantly faster to compute, due to fewer evaluations of Eq.~(\ref{eq:VertexSensitivity}) and high sparsity in Eq.~(\ref{eq:ConstraintDerivative}).

\begin{figure}
    \centering
    \includegraphics[trim={0cm 0.3cm 0cm 0.5cm},clip,width=0.75\linewidth]{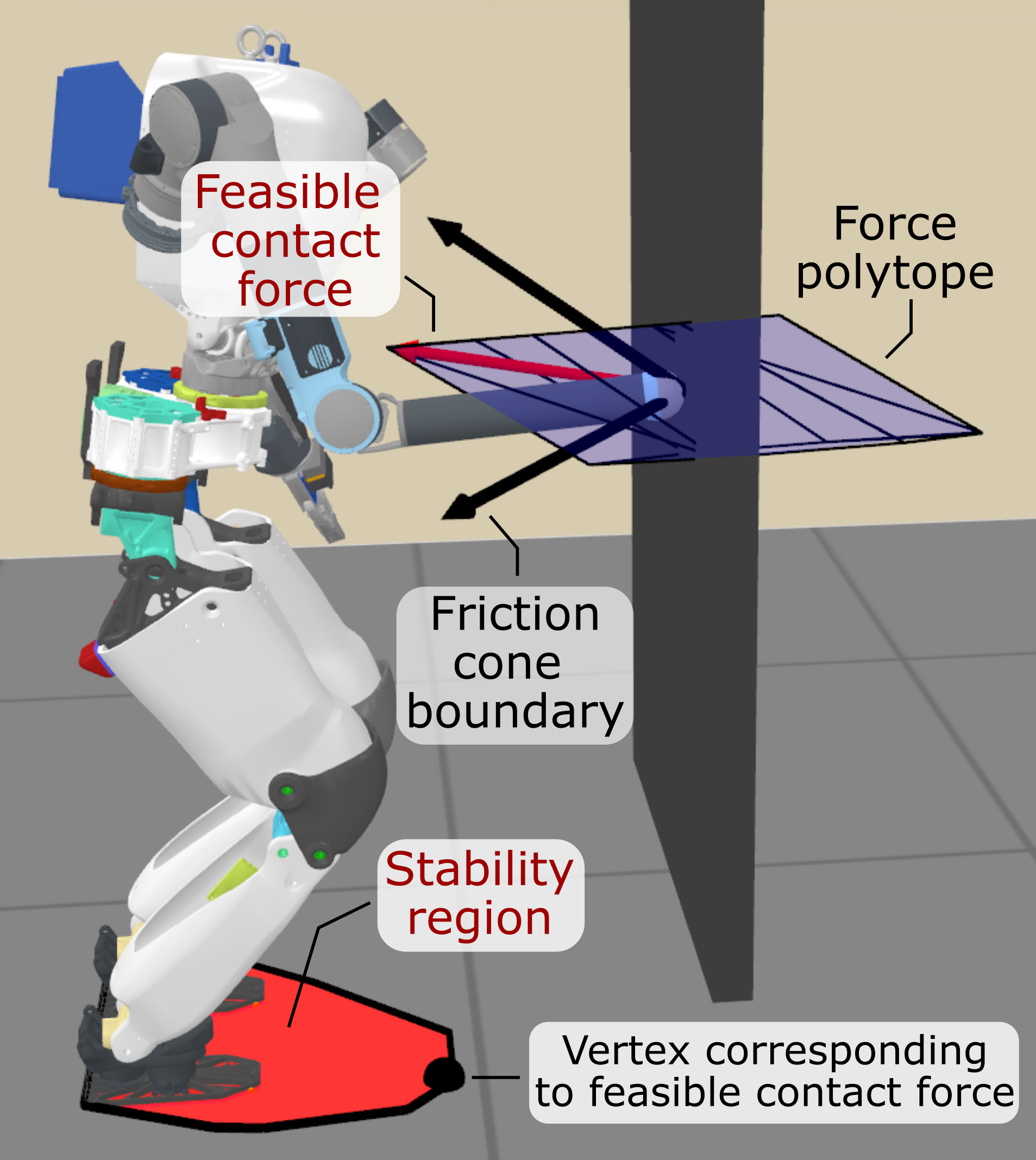}
    \caption{The 2d stability region encodes both actuation and friction constraints. When actuation constrained, the boundary of the stability region corresponds to the boundary of a force polytope. $\nabla m(\mathbf{q})$ is therefore a whole-body motion which rotates the force polytope and allows for increased contact force.}
    \label{fig:ConstraintSchematic}
\end{figure}

\subsection{Postural Sensitivity}
\revText{When the robot is bracing with the upper body (Fig.~\ref{fig:title_figure}), the kinematic tasks of the robot can be decomposed as}
\begin{equation}
\overbrace{
\begin{bmatrix}
\mathbf{J}_{h}\\
\mathbf{J}_{l}
\end{bmatrix}}^{\mathbf{J}_t} \mathbf{v} = 
\overbrace{
\begin{bmatrix}
\mathbf{p}_{h}\\
\mathbf{p}_{l}
\end{bmatrix}}^{\mathbf{p}_t},
\end{equation}
where $\mathbf{J}_{h}\mathbf{v}=\mathbf{p}_{h}$ encodes the high \revText{priority} tasks of \revText{active} contact constraints, grasping pose and desired CoM $xy$ position. The lower \revText{priority} task $\mathbf{J}_{l}\mathbf{v}=\mathbf{p}_{l}$ encodes a \textit{preferred posture} in the null space of $\mathbf{J}_{h}$ to maintain default chest and pelvis orientations along with a user-specified CoM height and bracing hand orientation. We are interested in modifying this preferred posture to improve stability, and refer to the magnitude of the projected gradient as the \textit{postural sensitivity}
\begin{equation}
s_q := || \mathbf{N}_{h} \,\nabla a^\ast_i (\mathbf{q}) ||^2,
\end{equation}
where $\mathbf{N}_{h}$ is the null space projector of $\mathbf{J}_{h}$ \cite{dietrich2015overview}. For cases such as standing on flat ground without hand contacts (often $s_q=0$) or in general the robot is primarily friction-constrained, the stability region cannot be modified through the posture. However for motions such as bracing against a wall and crawling (often $s_q>0$) or in general if the robot is actuation-constrained, it can be modified through posture. Mathematically, this occurs when elements of the joint torque margins $\mathbf{m}^{\pm}_\tau$ are $0$ in $\mathbf{x}^\ast$, meaning the actuation term actively constrains the stability region boundary. From the complementary slackness condition \cite{chvatal1983linear}, $\mathbf{y}^\ast_j>0$ for $j$ corresponding to each active actuation constraint and therefore we expect non-zero sensitivity from inspection of Eq. (\ref{eq:ConstraintDerivative}).
\section{Stability-Aware Teleoperation}\label{sec:Teleop}
We apply the stability analysis from the previous section to teleoperating multi-contact tasks on our humanoid. We focus on manipulation tasks where one hand braces against the environment to extend the reachable workspace of the robot, as shown in Fig.~\ref{fig:title_figure}. The humanoid has a hydraulically actuated lower body and electrically actuated upper body. As the electric joints have a lower torque capacity than the hydraulic ones, the robot is often actuation-constrained by the upper body when loading the arms. Put differently, the robot is likely to enter a posture sensitive state ($s_q > 0$) when loading the upper body, and unlikely to when only the legs are loaded. Therefore, we aim to locally optimize the teleoperation setpoints in order to improve the stability margin when the upper body is loaded.
\subsection{Teleoperation Framework} 
Teleoperation is performed by capturing user's motion with a Valve Index VR headset and two handheld controllers. Our framework tracks the operator's hand motions, with automatic retargeting of posture and contact objectives to optimize the stability margin $m$. As in \cite{bertrand_high-speed_2024}, an Inverse Kinematics (IK) module continuously computes a whole-body motion given the user's VR input, which in turn is tracked by a whole-body controller. The robot's motion is controlled by three operator inputs: (1) desired hand poses from the VR controllers, (2) a desired CoM position which the operator can move by joystick and (3) a desired hand action to toggle between load/unload (bracing hand) or open/close (grasping hand). The operator manually triggers when the bracing hand is loaded/unloaded, \revText{with a safety check that the robot's CoM lies within the support polygon of the feet to allow for unloading.} The IK module solves the optimization:
\begin{equation}
\begin{aligned}
 \label{eq:IKQP}
    \min_{\mathbf{v}_d} \quad   & ||\mathbf{J}_t \mathbf{v}_d - \mathbf{p}_t||^2_\mathbf{W} \\
    \textrm{s.t.} \quad         & \mathbf{v}_{min} \leq \mathbf{v}_d \leq \mathbf{v}_{max} \\
     \quad                       & \mathbf{A}_{xy}\mathbf{v}_d \leq \mathbf{h}(m_{min})
\end{aligned}
\end{equation}
where $\mathbf{W}$ is a weight matrix and $\mathbf{v}_d$ is the desired whole-body velocity, which is integrated each tick to compute a desired configuration $\mathbf{q}_d$. The bounds $\mathbf{v}_{min},\mathbf{v}_{max}$ prevents excessively rapid motion and enforce joint limits. $\mathbf{A}_{xy}$ is the linear $xy$ component of the centroidal momentum matrix \cite{orin_centroidal_2008} and $\mathbf{h}$ encodes a stability region constraint, such that the CoM remains within the stability region to a desired tolerance $m_{min}=4$cm. This safety-tolerance constraint is analogous to the 3d centroidal acceleration margins in Audren et al. \cite{audren_3-d_2018}.\startRev
\subsection{Retargeting Strategy}\label{sec:retargeting_strategy}
We leverage motion retargeting, the general process of mapping operator motion to robot motion \cite{rakita2019shared, darvish_teleoperation_2023}, to optimize the robot's stability. In general, the stability margin $m$ can be improved in two ways: moving the CoM away from the stability boundary and moving the stability boundary away from the CoM. The main idea of our framework is to rely primarily on the latter, preemptively growing the stability region to avoid excessive restriction of the robot's CoM. Notably, most prior works have used the former approach as a hard constraint \cite{orsolino_feasible_2020, audren_3-d_2018, samadi_balance_2019, roux_control_2021}. This is not preferable as it may interfere with the \textit{high-level} manipulation task. Instead, our approach is to improve stability through the \textit{lower-level} non-manipulation tasks and only impose a hard constraint on the CoM for critically low stability $m_{min}$. The retargeting objective is therefore: given \textit{some} feasible CoM setpoint, grow the stability region through retargeting contact position and posture, if possible. Given this, we opted for a straightforward input scheme (Tab.~\ref{table:retarget_scheme}) where the operator directly pilots the CoM, relying on the hard constraint $m\geq m_{min}$ to enforce a minimum tolerance. \stopRev
\begin{table}[hbt]
\caption{Teleoperation Objectives}
\centering
\label{table:retarget_scheme}
\begin{tabular}{ |c|c|c| } 
 \hline
 Objective & Nominal setpoint & Retargeting  \\
 \hline
 \hline
Bracing hand pos. & VR controller pos. & Pre-contact \\
 \hline
Bracing hand or. & VR controller or. & Post-contact \\
 \hline
Grasping hand pose & VR controller pose & None \\
 \hline
CoM $xy$ & VR controller joystick & $m_{min}$ tol. constraint \\
 \hline
CoM $z$ & VR controller joystick & Post-contact \\
 \hline
Chest orientation & Fixed & Post-contact \\
 \hline
Pelvis orientation & Fixed & Post-contact \\
 \hline
\end{tabular}
\end{table}
\subsection{Contact Retargeting}
We adopt a heuristic approach to retarget hand contacts by previewing the stability region and adjusting the hand position before contact is actually made. This is to maintain a purely local optimization which does not require extra steps such as re-establishing or sliding the hand after initial contact is made. The stability region preview is computed when the hand is moved within 1 meter of a candidate bracing surface and the hand's current position is used as preview contact point. The desired hand position $\mathbf{p}_{h}$ is then retargeted based on the gradient of the stability region's area $A$
\begin{equation}\label{eq:contactRetarget}
    \dot{\mathbf{p}}_h = k_c (\mathbf{I} - \mathbf{n}\mathbf{n}^T) \nabla A (\mathbf{p}_k),
\end{equation}
where $\mathbf{n}$ is the normal of the candidate surface and $k_c$ is a scalar gain. The hand contact is only retargeted along the plane of the surface, such that the user maintains control of the distance from the hand to the surface. A maximum hand contact adjustment is enforced to prevent excessive retargeting.
\subsection{Posture Retargeting}\label{sec:posture_retargeting}
In contrast, posture retargeting is performed on the gradient of the stability margin after a hand contact has been established. This retargets the bracing hand orientation, chest, pelvis, and CoM height setpoints. For posture objective $i$ with task Jacobian $\mathbf{J}_i$, we retarget task setpoint $\mathbf{t}_i$ based on the stability conditions of the robot.
\begin{equation}\label{eq:postureRetarget}
\begin{aligned}
	\dot{\mathbf{t}}_i &= 
	\begin{cases}
        \, \dot{t}_0 (\Delta\mathbf{\hat{t}}_{i,nom}) & \text{if} \hspace{0.2cm} s_q = 0 \,\vee \, m > m^+ \hspace{0.2cm} (a) \\[2pt]
        \, k_q \mathbf{J}_i \mathbf{N}_{h} \nabla m(\mathbf{q}) & \text{else if} \hspace{0.2cm} s_q > s_q^+ \hspace{0.92cm} (b) \\[2pt]
        \, \mathbf{0} & \text{else} \hspace{2.53cm} (c) \\
    \end{cases}
\end{aligned}
\end{equation}

\textit{Nominal teleoperation} (\ref{eq:postureRetarget}a) is performed when the postural sensitivity is zero or (optionally) when the robot has a sufficiently large stability margin $m^+=15$cm. Under these conditions, the user regains full control of all posture objectives as they do not directly affect the stability margin. In this case, the objective resets to nominal at a fixed rate $\dot{t}_0$ based on the error from nominal $\Delta\mathbf{t}_{i,nom} = \mathbf{t}_{i,nom} - \mathbf{t}_{i}$. The threshold $m^+$ optionally provides the user with full control when stability may not be critical.

\textit{Posture retargeting} (\ref{eq:postureRetarget}b) is performed when the robot has high postural sensitivity ($s>s_q^+$). While the condition $s_q>0$ could also be used, we found retargeting under low postural sensitivity to cause undesirable behavior for two reasons. Firstly, the robot performs large motions to achieve minimal increases in stability margin, which is unwanted. Additionally the accuracy of the gradient is poor for low sensitivity values, as the sensitivity analysis only captures motion perpendicular to the region. In such cases the tangent motion is often non-negligible and causes the gradient to become inaccurate. We use the value $s_q^+ = 0.01$, which corresponds to approximately 10$\si{\degree}$ joint motion per 1cm improvement of stability margin.
\begin{figure}
    \centering
    \includegraphics[trim={1.0cm 0.5cm 0.0cm 0cm},clip,width=\columnwidth]{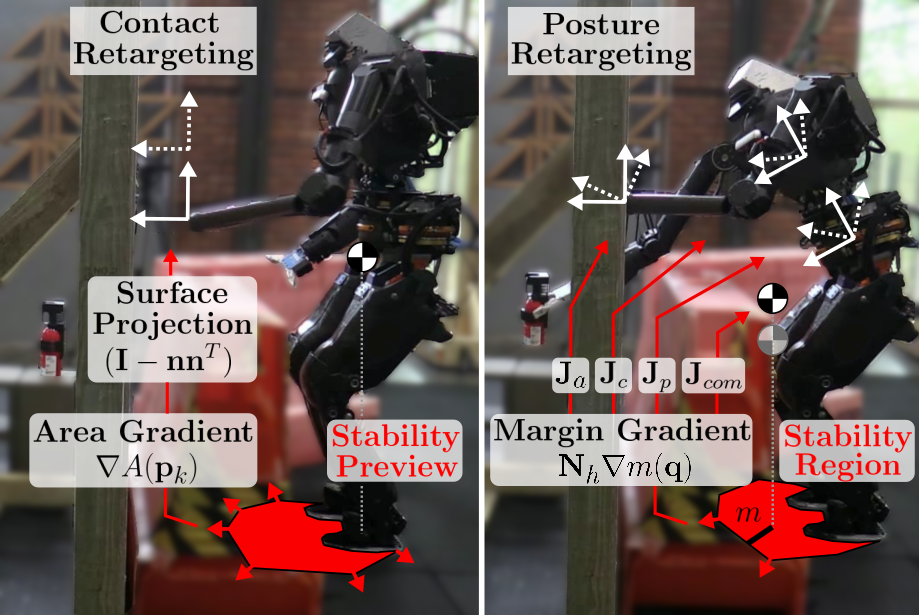}
    \caption{Control flow for contact and posture retargeting. The contacting hand is adjusted prior to loading the hand by previewing the stability region at the robot's current configuration. The area gradient is used to grow the preview region. When the \textit{posture sensitivity} $s_q$ is sufficiently high, the stability margin gradient informs the adjustment of the posture setpoints.}
    \label{fig:OverheadStabilityObjective}
\end{figure}

A \textit{Freeze retargeting} (\ref{eq:postureRetarget}c) mode avoids unwanted state-switching between nominal and optimized postures when near the boundary of $s_q=0$.

Finally, two measures are implemented to avoid excessive posture adjustment. Firstly, joint limits are accounted in the calculation of $\nabla a^\ast_i(\mathbf{q})$ (Eq.~(\ref{eq:GenericGradienRelatedToTimeDerivative})) such that if joint $j$ is close to a limit and $\frac{\partial a^\ast_i}{\partial t}(\mathbf{v} = \hat{\mathbf{v}}_j)$ drives the joint further towards the joint limit, it is considered to have no postural sensitivity. Secondly, a maximum deviation from the nominal setpoint $\mathbf{t}_{i,nom}$ is imposed.
\section{Real-time Controller Integration}
\begin{figure}
    \centering
    \includegraphics[width=0.8\linewidth]{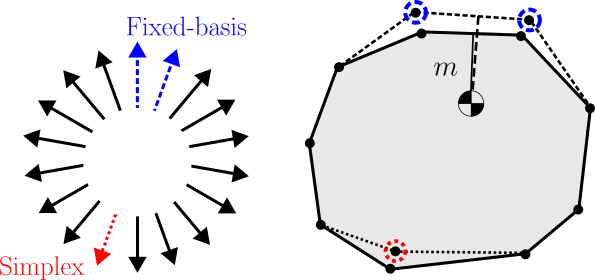}
    \caption{The stability region is incrementally updated using a set of fixed query directions, in which one full Linear Program solve (Eq.~(\ref{eq:CoMLP})) and two fast, fixed-basis solves (Eq.~(\ref{eq:fixedBasisUpdate})) are performed.}
    \label{fig:updatePolicy}
\end{figure}

Deploying the presented retargeting scheme to hardware required integration with a real-time controller process, in which the IK solver (Eq.~(\ref{eq:IKQP})) runs alongside a state estimator and whole-body controller. The first step towards achieving real-time stability assessment was the implementation of an allocation-free Java Linear Program solver, which is open source and available online\footnote{Package \texttt{linearProgram} at \url{www.github.com/ihmcrobotics/ihmc-convex-optimization}}. The other enabling factor is an efficient, incremental update rule for the stability region. Although the Iterative Projection (IP) algorithm \cite{bretl_testing_2008} is the canonical framework in motion planning for computing the stability region, its foundational premises do not apply well to fast, incremental updates for two reasons. Firstly IP solves to arbitrary precision assuming a non-linearized friction cone, in which the stability region boundary is smooth. Applying this to a linearized friction cone causes each constraint edge to be computed twice, with the second query ``confirming'' it is the outer bound. Therefore we instead choose to simply use a set of $n_q=18$ fixed, equally-spaced query directions. Secondly, IP is designed to solve the region from scratch, as opposed to our application in a kinematics solver in which the region changes gradually. Instead, we only perform one LP solve (Eq.~(\ref{eq:CoMLP})) each tick and maintain an accurate estimate of $m$ with a fast, fixed-basis update of the minimum-margin edge (Fig.~\ref{fig:updatePolicy})
%\cite{bertrand_high-speed_2024}.
\begin{table}[t]
\begin{center}
\caption{Stability Region Benchmark}\label{table:region_benchmark}
\begin{tabular}{ |c|c|c| } 
\hline
 & Avg. time & Max $m$ error \\
\hline
\hline
Iterative Projection & 1731$\mu\text{s}$ & 0mm \\
\hline
Fixed Queries, Full Update &  698$\mu\text{s}$ & 0.9mm \\
\hline
Fixed Queries, Incremental Update  & 79$\mu\text{s}$ & 3.5mm \\
\hline
\hline
Stability Margin Gradient ($\nabla m(\mathbf{q})$) & $334\mu\text{s}$ & - \\
\hline
\end{tabular}
\end{center}
\end{table}
\begin{figure*}
    \centering
    \includegraphics[trim={1.4cm 0 1.5cm 0},clip,width=\textwidth]{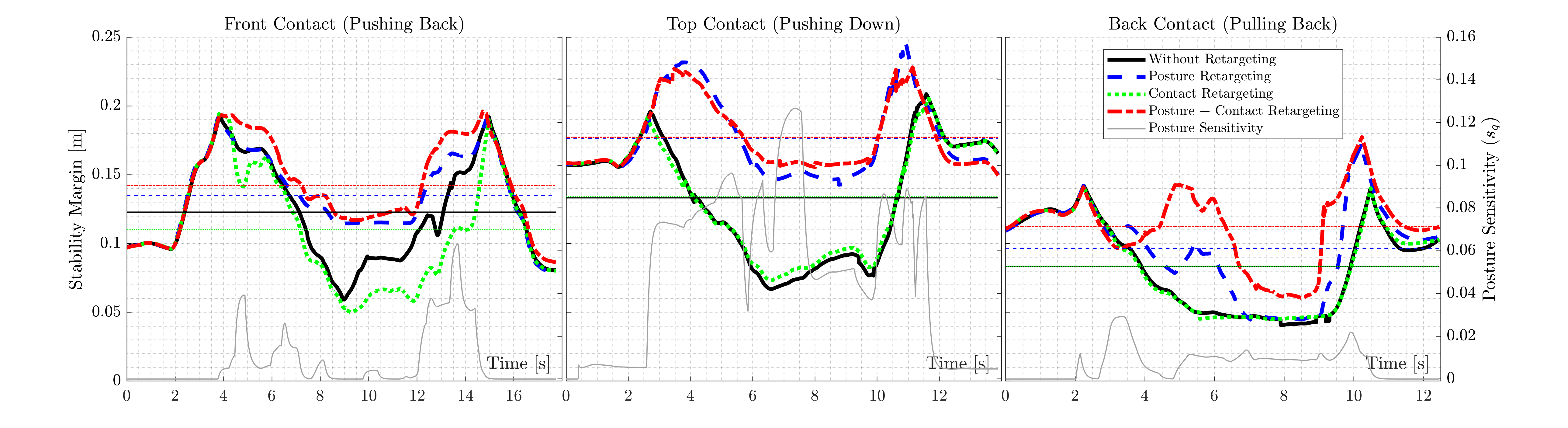}
    \caption{Ablation study showing the contribution of posture and contact retargeting in 3 simulated scenarios (Fig.~\ref{fig:simulation_cases}). The posture sensitivity (for non-retargeted case) and average stability margins are included for comparison in thin lines. Notably, contact-only retargeting performs poorly, posture-only retargeting is substantially higher than nominal, and contact+posture retargeting performs best in each case.}
    \label{fig:simulated_ablation_study}
\end{figure*}
\begin{equation}\label{eq:fixedBasisUpdate}
    \mathbf{x}_B = \mathbf{A}^{-1}_{B}(\mathbf{b} - \mathbf{A}_N\mathbf{x}_N),
\end{equation}
where $B$ and $N$ represent the basis and non-basis components of $\mathbf{A}$ and $\mathbf{x}$, respectively \cite{chvatal1983linear}. We verify our incremental update approach does not sacrifice accuracy, finding the deviation in computed stability margin to be negligible. Table \ref{table:region_benchmark} shows the benchmark timing data. Table \ref{table:region_benchmark} shows a timing comparison between the different update policies and the gradient computation.
\section{Results}
\subsection{Simulation}
\begin{figure}
    \centering
    \includegraphics[width=\linewidth]{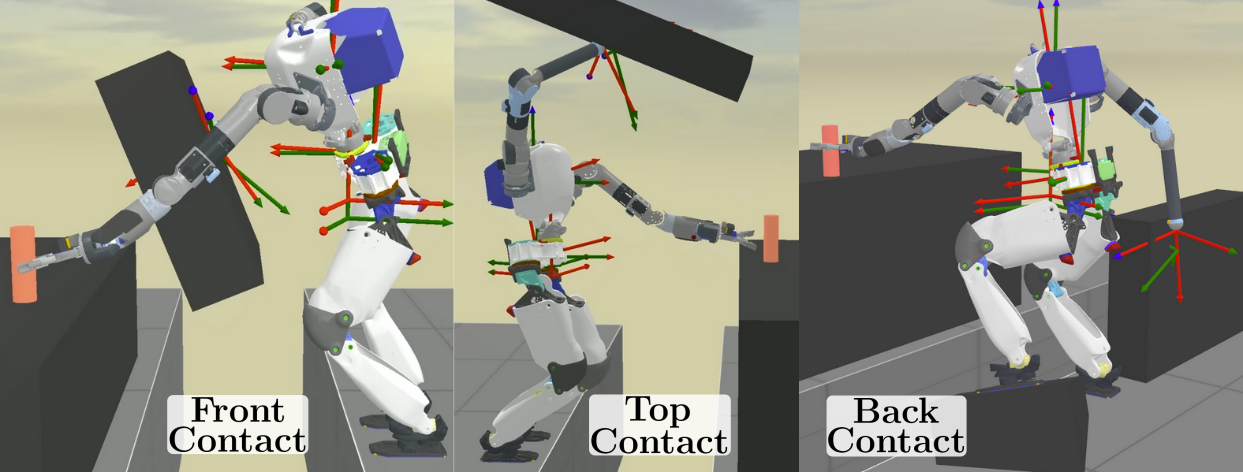}
    \caption{Three manipulation scenarios were tested in simulation. In each, the robot uses a hand contact to extend the reachable workspace of the left arm and grab the object. The nominal (green) and retargeted (red) setpoints for the pelvis, chest, CoM height and right hand are visualized.}
    \label{fig:simulation_cases}
\end{figure}
\begin{figure}[b]
    \centering
    \includegraphics[trim={1.1cm 0cm 1.6cm 0.0cm},clip,width=\linewidth]{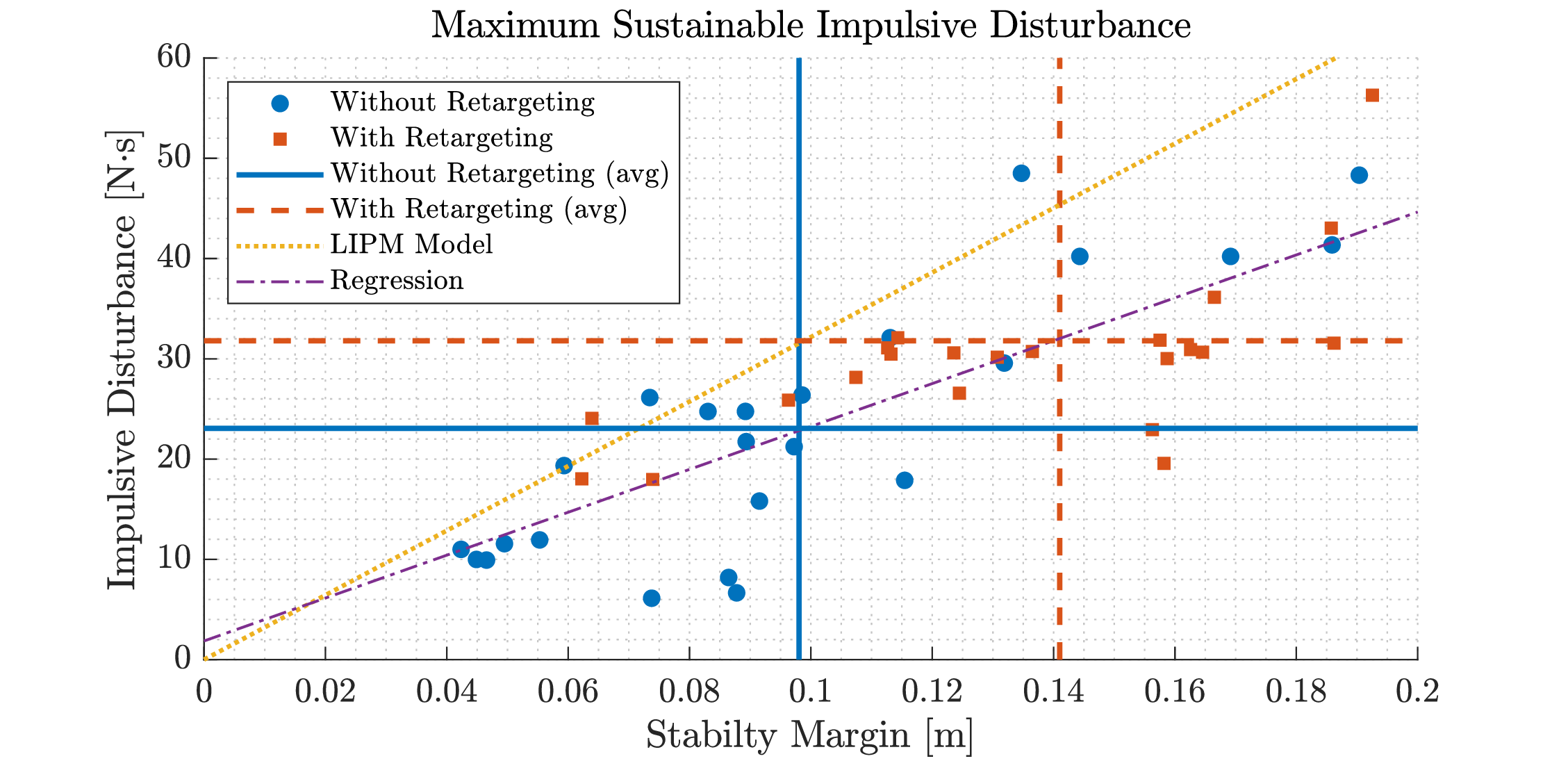}
    \caption{Simulation study to determine the correlation of the stability margin $m$ and the maximum sustainable impulse.}
    \label{fig:disturbance_rejection}
\end{figure}
We tested our framework by teleoperating a manipulation task under three distinct contact conditions. Shown in Fig.~\ref{fig:simulation_cases}, these cases are a pitched wall in front, a ceiling above and a wall behind the robot with uneven footholds (all unilateral contact). By varying contact conditions, we aim to test the versatility of the framework and understand the relative effectiveness of posture and contact retargeting. In each scenario, a baseline trajectory is generated without retargeting in which the operator's motion and inputs are logged. We then performed an ablation study by replaying the logged operator trajectory with posture and contact retargeting separately and both enabled. In all cases, the simulated grasp was evaluated relative to a predefined ``perfect grasp'' and deemed successful if the off-axis error did not exceed 5cm position and 20$\si{\degree}$ orientation. Posture retargeting was performed for any posture with sufficient sensitivity $s^+_q= 0.01$ (i.e. $m^+$ was disabled in (\ref{eq:postureRetarget}a)). The robot successfully retrieved the canister for all simulations, validating the grasping hand pose was not significantly modified by the retargeting.

Fig.~\ref{fig:simulated_ablation_study} shows the robot's stability for each of the simulated trajectories. These plots reflect the margin computed by the \textit{controller} using the robot's current configuration, which in general differs from the desired stability margin. At the start and end of each trajectory, the robot's CoM is near the middle of the feet and the robot is not in a posture-sensitive state. The peaks in stability correspond to the CoM crossing the middle of the region (arm slightly loaded), while the central dip is the robot heavily loading the arm.
\begin{figure}
    \centering
    \includegraphics[trim={0cm 0.1cm 0cm 0.1cm},clip,width=0.64\linewidth]{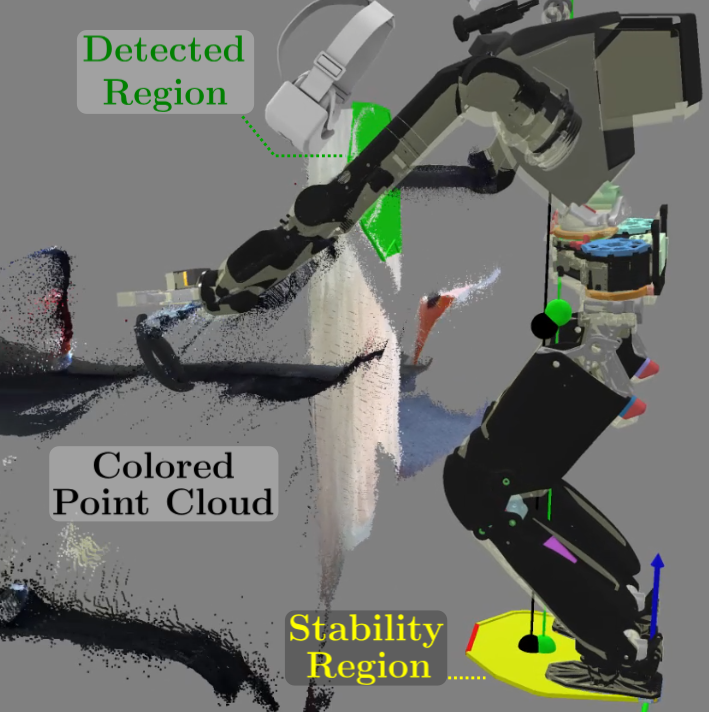}
    \caption{VR perspective displaying current and desired (transparent) robot state, current (black column) and desired (green column) CoM positions, and live perception data.}
    \label{fig:vr_perspective}
\end{figure}

We first note that, on average, contact-only retargeting actually decreases the stability by 3\% while posture-only retargeting increases by 20$\%$. This is somewhat expected, as the posture-based adjustment reactively improves stability and the contact adjustment heuristically grows the preview region.\blockcomment{In the first scenario, the contact-only trajectory happens to modify the upper-body Jacobian in a way that reduces stability, despite the hand contact otherwise being improved. However, when enabling both posture and contact retargeting this is corrected and the optimized contact position results in 16\% higher stability.} On average, for combined posture and contact retargeting there is a 27\% stability increase. Mathematically, the contact adjustment heuristically aims to keep the moment balance constraint of Eq.~(\ref{eq:CoMLP}) \textit{out} of the optimal basis, increasing the likelihood that actuation constraints are \textit{in} the optimal basis and can in turn be relaxed through posture adjustment. Overall, this heuristic is validated by scenarios 1 and 3, while in scenario 2 no significant change is observed.

To evaluate the centroidal stability margin $m$ as a robustness indicator, we determine the maximum sustainable impulsive disturbance during sampled times in the three above trajectories, shown in Fig.~\ref{fig:disturbance_rejection}. For each trajectory, 8 times are sampled in the middle 50\% and a maximum sustainable impulse is determined through binary search. The impulse is applied to the pelvis in the $xy$ plane and pushes the robot towards the minimum-margin edge (dotted line in Fig.~\ref{fig:OverheadStability}). A regression ($R^2=0.69$) validates the correlation of stability margin with impulse magnitude, and we attribute the variation from linearity to the actuation constraint (Eq.~(\ref{eq:CoMLP})) being a local model. % For reference, the LIPM-based Capture Point model \cite{englsberger2011bipedal} is also plotted.
\subsection{Hardware}
\begin{figure}
    \centering
    \includegraphics[trim={0.65cm 0 0cm 0},clip,width=0.9\linewidth]{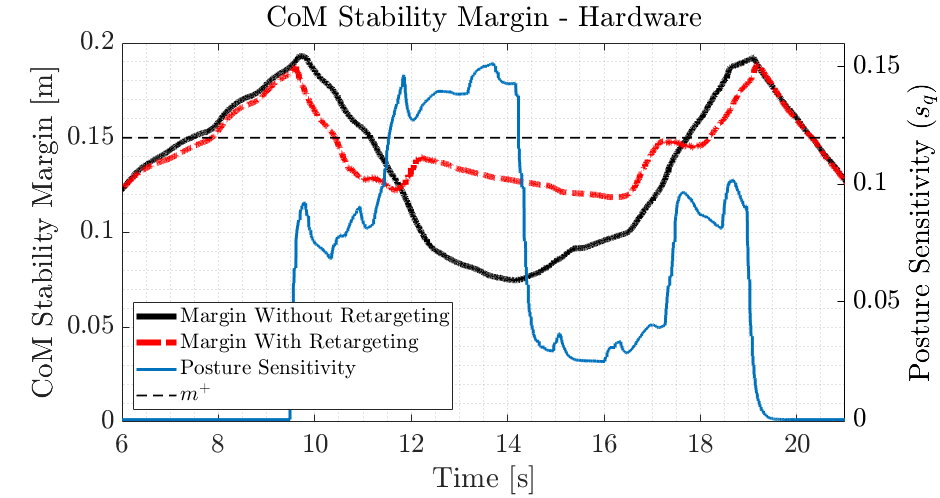}
    \caption{Stability margin $m$ during a hardware teleoperation task. The posture sensitivity $s_q$ highlights when retargeting was activated.}
    \label{fig:hardware_margin}
\end{figure}

Our humanoid is equipped with an Intel Realsense D455 Depth camera and ZED 2 Stereo camera which perceive the environment in front of the robot. The colored point cloud data from both sensors are displayed in the operator's VR perspective, as shown in Fig.~\ref{fig:vr_perspective}. Additionally, we extract planar regions to detect candidate bracing surfaces, using the region closest to the hand (if within 1m). % \cite{mishra2021gpu} ommitted
We teleoperated three trajectories to retrieve a canister, varying the wall orientation and canister position. As in the simulated experiments, the operator's inputs were recorded and replayed to evaluate the retargeting. We observed an average stability margin increase of 7\% and average maximum stability margin increase of 121\%. Fig.~\ref{fig:hardware_margin} shows the stability margin for one of the teleoperated trajectories, with the posture sensitivity $s_q$ included. The portion of the trajectory where $s_q > s_q^+ = 0.01$ and $m < m^+ =0.15$ designates when the upper body is heavily loaded and the posture retargeting is activated. As with the simulated trajectories, the initial contact adjustment actually decreases the stability, but is corrected when the posture adjustment is activated.

\begin{figure}
    \centering
  \includegraphics[trim={1.25cm 0 2cm 0},clip,width=1\linewidth]{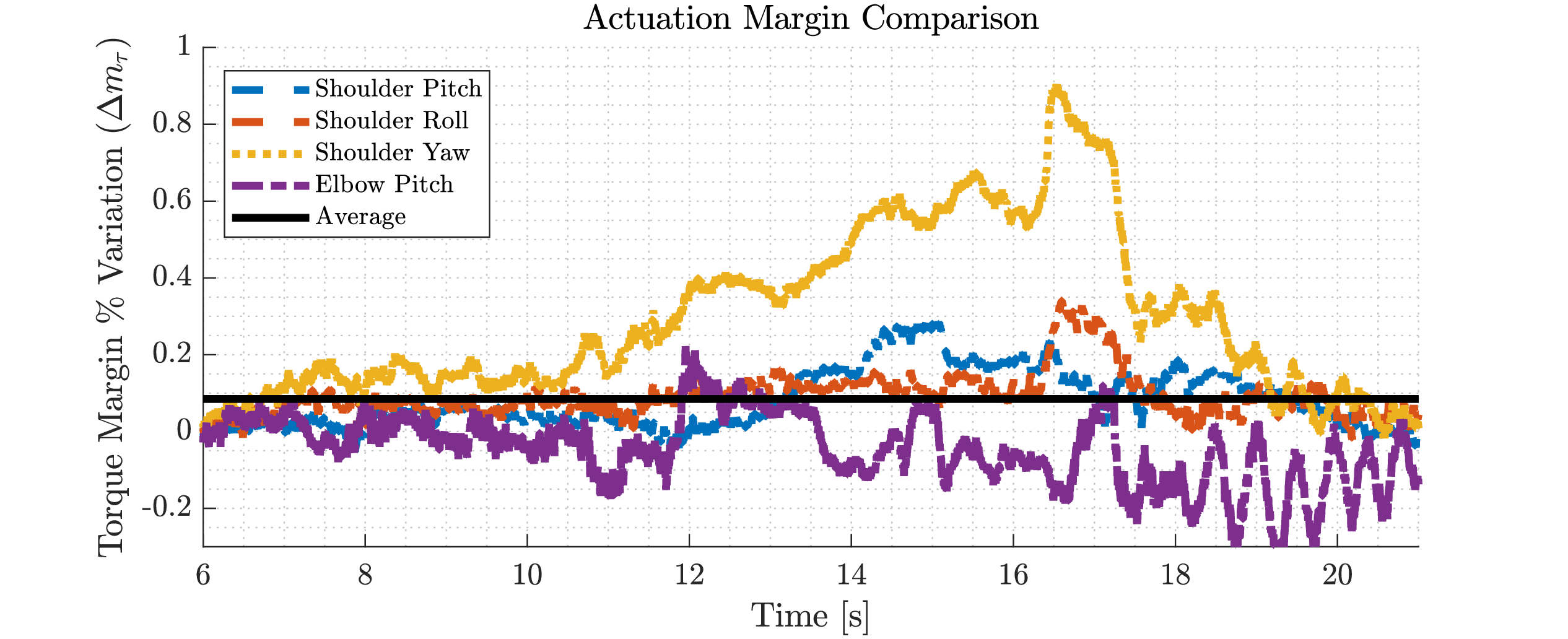}
    \caption{Torque limit saturation comparison, in which a $10\%$ increase in joint torque margin $m_{\tau}$ was observed.}
    \label{fig:hardware_torque_margin}
\end{figure}

We also observed the joint margin $m_{\tau}$, given by
\begin{equation}
    m_{\tau} = 1 -  \left | \frac{\tau_{meas}}{\tau_{max}} \right |,
\end{equation}
where $\tau_{meas}$, $\tau_{max}$ are the measured and maximum torques. We found that on average, $m_{\tau}$ increased by $12.6\%$ ($10.0\%$ in Fig.~\ref{fig:hardware_torque_margin}) with retargeting. Note this additional torque margin will only be present when it improves the robot's stability and will not when the robot is friction constrained, for example. This differs from similar approaches which directly penalize joint torques and may unnecessarily impact the performance of other objectives.
\section{Conclusion}
Stable and versatile teleoperation of humanoid robots plays an important role in fully utilizing these systems. Particularly as learning-based approaches increasingly use teleoperation data \cite{he2024learning,gu_humanoid_2025}, it is vital that teleoperation systems are designed for the kind of complex terrains in which humanoids are meant to be operated. Towards this goal, we designed a teleoperation system with hand contacts and actuation limits as a primary focus. Through an analytical calculation of the stability region gradient, we formed a teleoperation scheme which improves stability by retargeting the robot's contact point and posture. We demonstrated that our framework successfully increases impulse resilience by testing in simulation for highly irregular multi-contact scenarios along with conducting hardware experiments.

The presented approach has four notable limitations based on simplifying assumptions: (i) a ``snapshot'' actuation model (Eq.~(\ref{eq:ActuationConstraint})), (ii) cancellation of sensitivity near joint limits, (iii) a quasi-static model, (iv) the heuristic that restricting CoM motion will impact task performance (\ref{sec:retargeting_strategy}). \revText{Assumption (i) is addressed in \cite{ferrolho2023roloma} through} a trajectory optimization which considers robustness along with whole-body state, however solve times are $\approx30$s and not real-time compatible. For (ii), while mature whole-body approaches such as \cite{ferrolho2023roloma, corberes2025perceptive} do consider kinematic limits in the \textit{nominal} plan, like ours, they do not in the model of \textit{robustness} beyond direct penalty on limit proximity. \revText{Assumption} (iii) is not an inherent limitation, provided a reference acceleration is available \cite{orsolino_feasible_2020}. An extension of this work could investigate local refinement of a nominal dynamic plan, towards increasing robustness, using the same sensitivity analysis. Finally, (iv) requires relating the robot's reachable workspace with centroidal motion. This would enable the retargeting of the robot's CoM in scenarios where task performance is not impacted.

\appendix
\textit{Gravitational Torque Contribution:} The configuration $\mathbf{q}$ parameterizes both the constraint matrix $\mathbf{A}$ (from $\mathbf{J}_c(\mathbf{q})$) and constraint vector $\mathbf{b}$ (from $\mathbf{g}(\mathbf{q})$). We compared the relative contribution of these two terms by simulating random contact conditions and observing the stability variation $\delta m$ resulting from a postural variation $\delta \mathbf{q}$ (projected by $\mathbf{N}_{h}$). This simulation was then rerun with $\mathbf{A}$ fixed and again with $\mathbf{b}$ fixed, at 1000 random configurations. We observed normalized errors of 7.0$\%$ for fixed $\mathbf{A}$ and $88.9\%$ for fixed $\mathbf{b}$, relative to the true margin variation. We conclude a sensitivity analysis with respect to only $\mathbf{A}$ captures the majority of postural dependence.

% \renewcommand{\arraystretch}{1.0}
% \begin{table}[h!]
% \begin{center}
% \caption{Average Error under Postural Variation}\label{table:grav_torques}
% \begin{tabular}{ |c| } 

%  \hline
%    Error $\mathbf{A}$ fixed:\,\, $|\delta m - \delta m_A|$ = $0.930|\delta m|$ \\[1pt]
%     \hline
%    Error $\mathbf{b}$ fixed:\,\, $|\delta m - \delta m_b|$ = $0.111|\delta m|$ \\[1pt]
%  \hline
% \end{tabular}
% \end{center}
% \end{table}
%
  % \section*{References}
  % Generated by IEEEtran.bst, version: 1.14 (2015/08/26)


\begin{thebibliography}{10}
\providecommand{\url}[1]{#1}
\csname url@samestyle\endcsname
\providecommand{\newblock}{\relax}
\providecommand{\bibinfo}[2]{#2}
\providecommand{\BIBentrySTDinterwordspacing}{\spaceskip=0pt\relax}
\providecommand{\BIBentryALTinterwordstretchfactor}{4}
\providecommand{\BIBentryALTinterwordspacing}{\spaceskip=\fontdimen2\font plus
\BIBentryALTinterwordstretchfactor\fontdimen3\font minus \fontdimen4\font\relax}
\providecommand{\BIBforeignlanguage}[2]{{%
\expandafter\ifx\csname l@#1\endcsname\relax
\typeout{** WARNING: IEEEtran.bst: No hyphenation pattern has been}%
\typeout{** loaded for the language `#1'. Using the pattern for}%
\typeout{** the default language instead.}%
\else
\language=\csname l@#1\endcsname
\fi
#2}}
\providecommand{\BIBdecl}{\relax}
\BIBdecl

\bibitem{gu_humanoid_2025}
\BIBentryALTinterwordspacing
Z.~Gu, J.~Li, W.~Shen, W.~Yu, Z.~Xie, S.~McCrory, X.~Cheng, A.~Shamsah, R.~Griffin, C.~K. Liu, A.~Kheddar, X.~B. Peng, Y.~Zhu, G.~Shi, Q.~Nguyen, G.~Cheng, H.~Gao, and Y.~Zhao., ``Humanoid locomotion and manipulation: Current progress and challenges in control, planning, and learning,'' \emph{IEEE/ASME Transactions on Mechatronics. DOI: 10.1109/TMECH.2025.3579247}, 2025.
\BIBentrySTDinterwordspacing

\bibitem{bretl_testing_2008}
\BIBentryALTinterwordspacing
T.~Bretl and S.~Lall, ``Testing static equilibrium for legged robots,'' vol.~24, no.~4, pp. 794--807.
\BIBentrySTDinterwordspacing

\bibitem{orsolino_feasible_2020}
\BIBentryALTinterwordspacing
R.~Orsolino, M.~Focchi, S.~Caron, G.~Raiola, V.~Barasuol, D.~G. Caldwell, and C.~Semini, ``Feasible region: An actuation-aware extension of the support region,'' vol.~36, no.~4, pp. 1239--1255.
\BIBentrySTDinterwordspacing

\bibitem{selvaggio2021autonomy}
M.~Selvaggio, M.~Cognetti, S.~Nikolaidis, S.~Ivaldi, and B.~Siciliano, ``Autonomy in physical human-robot interaction: A brief survey,'' \emph{IEEE Robotics and Automation Letters}, vol.~6, no.~4, pp. 7989--7996, 2021.

\bibitem{englsberger2011bipedal}
J.~Englsberger, C.~Ott, M.~A. Roa, A.~Albu-Sch{\"a}ffer, and G.~Hirzinger, ``Bipedal walking control based on capture point dynamics,'' in \emph{2011 IEEE/RSJ international conference on intelligent robots and systems}.\hskip 1em plus 0.5em minus 0.4em\relax IEEE, 2011, pp. 4420--4427.

\bibitem{dai_planning_2016}
\BIBentryALTinterwordspacing
H.~Dai and R.~Tedrake, ``Planning robust walking motion on uneven terrain via convex optimization,'' in \emph{2016 {IEEE}-{RAS} 16th International Conference on Humanoid Robots (Humanoids)}.\hskip 1em plus 0.5em minus 0.4em\relax {IEEE}, pp. 579--586.
\BIBentrySTDinterwordspacing

\bibitem{fernbach2020c}
P.~Fernbach, S.~Tonneau, O.~Stasse, J.~Carpentier, and M.~Taïx, ``C-croc: Continuous and convex resolution of centroidal dynamic trajectories for legged robots in multicontact scenarios,'' \emph{IEEE Transactions on Robotics}, vol.~36, no.~3, pp. 676--691, 2020.

\bibitem{wang2024online}
J.~Wang, S.~Kim, T.~S. Lembono, W.~Du, J.~Shim, S.~Samadi, K.~Wang, V.~Ivan, S.~Calinon, S.~Vijayakumar \emph{et~al.}, ``Online multicontact receding horizon planning via value function approximation,'' \emph{IEEE Transactions on Robotics}, vol.~40, pp. 2791--2810, 2024.

\bibitem{del_prete_fast_2016}
\BIBentryALTinterwordspacing
A.~Del~Prete, S.~Tonneau, and N.~Mansard, ``Fast algorithms to test robust static equilibrium for legged robots,'' in \emph{2016 {IEEE} International Conference on Robotics and Automation ({ICRA})}.\hskip 1em plus 0.5em minus 0.4em\relax {IEEE}, pp. 1601--1607.
\BIBentrySTDinterwordspacing

\bibitem{escande_planning_2013}
\BIBentryALTinterwordspacing
A.~Escande, A.~Kheddar, and S.~Miossec, ``Planning contact points for humanoid robots,'' vol.~61, no.~5, pp. 428--442.
\BIBentrySTDinterwordspacing

\bibitem{nozawa_three-dimensional_2016}
\BIBentryALTinterwordspacing
S.~Nozawa, M.~Kanazawa, Y.~Kakiuchi, K.~Okada, T.~Yoshiike, and M.~Inaba, ``Three-dimensional humanoid motion planning using {COM} feasible region and its application to ladder climbing tasks,'' in \emph{2016 {IEEE}-{RAS} 16th International Conference on Humanoid Robots (Humanoids)}.\hskip 1em plus 0.5em minus 0.4em\relax {IEEE}, pp. 49--56.
\BIBentrySTDinterwordspacing

\bibitem{audren_3-d_2018}
\BIBentryALTinterwordspacing
H.~Audren and A.~Kheddar, ``3-d robust stability polyhedron in multicontact,'' vol.~34, no.~2, pp. 388--403. [Online]. Available: \url{http://ieeexplore.ieee.org/document/8289420/}
\BIBentrySTDinterwordspacing

\bibitem{del_prete_zero_2018}
\BIBentryALTinterwordspacing
A.~Del~Prete, S.~Tonneau, and N.~Mansard, ``Zero step capturability for legged robots in multicontact,'' vol.~34, no.~4, pp. 1021--1034.
\BIBentrySTDinterwordspacing

\bibitem{ferrolho2023roloma}
H.~Ferrolho, V.~Ivan, W.~Merkt, I.~Havoutis, and S.~Vijayakumar, ``Roloma: Robust loco-manipulation for quadruped robots with arms,'' \emph{Autonomous Robots}, vol.~47, no.~8, pp. 1463--1481, 2023.

\bibitem{samadi_balance_2019}
S.~Samadi, S.~Caron, A.~Tanguy, and A.~Kheddar, ``Balance of humanoid robots in a mix of fixed and sliding multi-contact scenarios,'' in \emph{2020 IEEE International Conference on Robotics and Automation (ICRA)}, 2020, pp. 6590--6596.

\bibitem{roux_control_2021}
\BIBentryALTinterwordspacing
J.~Roux, S.~Samadi, E.~Kuroiwa, T.~Yoshiike, and A.~Kheddar, ``Control of humanoid in multiple fixed and moving unilateral contacts,'' in \emph{2021 20th International Conference on Advanced Robotics ({ICAR})}.\hskip 1em plus 0.5em minus 0.4em\relax {IEEE}, pp. 793--799.
\BIBentrySTDinterwordspacing

\bibitem{hirukawa_universal_2006}
\BIBentryALTinterwordspacing
H.~Hirukawa, S.~Hattori, K.~Harada, S.~Kajita, K.~Kaneko, F.~Kanehiro, K.~Fujiwara, and M.~Morisawa, ``A universal stability criterion of the foot contact of legged robots - adios {ZMP},'' in \emph{Proceedings 2006 {IEEE} International Conference on Robotics and Automation, 2006. {ICRA} 2006.}\hskip 1em plus 0.5em minus 0.4em\relax {IEEE}, pp. 1976--1983.
\BIBentrySTDinterwordspacing

\bibitem{caron_leveraging_2015}
\BIBentryALTinterwordspacing
S.~Caron, Q.~Cuong~Pham, and Y.~Nakamura, ``Leveraging cone double description for multi-contact stability of humanoids with applications to statics and dynamics,'' in \emph{Robotics: Science and Systems {XI}}.\hskip 1em plus 0.5em minus 0.4em\relax Robotics: Science and Systems Foundation.
\BIBentrySTDinterwordspacing

\bibitem{caron_zmp_2017}
\BIBentryALTinterwordspacing
S.~Caron, Q.-C. Pham, and Y.~Nakamura, ``{ZMP} support areas for multicontact mobility under frictional constraints,'' vol.~33, no.~1, pp. 67--80. [Online]. Available: \url{http://ieeexplore.ieee.org/document/7782743/}
\BIBentrySTDinterwordspacing

\bibitem{murooka_centroidal_2022}
\BIBentryALTinterwordspacing
M.~Murooka, M.~Morisawa, and F.~Kanehiro, ``Centroidal trajectory generation and stabilization based on preview control for humanoid multi-contact motion,'' vol.~7, no.~3, pp. 8225--8232. [Online]. Available: \url{https://ieeexplore.ieee.org/document/9807369/}
\BIBentrySTDinterwordspacing

\bibitem{rouxel_multicontact_2022}
\BIBentryALTinterwordspacing
Q.~Rouxel, K.~Yuan, R.~Wen, and Z.~Li, ``Multicontact motion retargeting using whole-body optimization of full kinematics and sequential force equilibrium,'' vol.~27, no.~5, pp. 4188--4198. [Online]. Available: \url{https://ieeexplore.ieee.org/document/9728754/}
\BIBentrySTDinterwordspacing

\bibitem{darvish_teleoperation_2023}
\BIBentryALTinterwordspacing
K.~Darvish, L.~Penco, J.~Ramos, R.~Cisneros, J.~Pratt, E.~Yoshida, S.~Ivaldi, and D.~Pucci, ``Teleoperation of humanoid robots: A survey,'' vol.~39, no.~3, pp. 1706--1727. [Online]. Available: \url{https://ieeexplore.ieee.org/document/10035484/}
\BIBentrySTDinterwordspacing

\bibitem{ramos2018dynamic}
J.~Ramos and S.~Kim, ``Humanoid dynamic synchronization through whole-body bilateral feedback teleoperation,'' \emph{IEEE Transactions on Robotics}, vol.~34, no.~4, pp. 953--965, 2018.

\bibitem{he2024learning}
T.~He, Z.~Luo, W.~Xiao, C.~Zhang, K.~Kitani, C.~Liu, and G.~Shi, ``Learning human-to-humanoid real-time whole-body teleoperation,'' in \emph{2024 IEEE/RSJ International Conference on Intelligent Robots and Systems (IROS)}.\hskip 1em plus 0.5em minus 0.4em\relax IEEE, 2024, pp. 8944--8951.

\bibitem{grandia_doc_2023}
\BIBentryALTinterwordspacing
R.~Grandia, F.~Farshidian, E.~Knoop, C.~Schumacher, M.~Hutter, and M.~Bächer, ``{DOC}: Differentiable optimal control for retargeting motions onto legged robots,'' vol.~42, no.~4, pp. 1--14.
\BIBentrySTDinterwordspacing

\bibitem{losey2018review}
D.~P. Losey, C.~G. McDonald, E.~Battaglia, and M.~K. O'Malley, ``A review of intent detection, arbitration, and communication aspects of shared control for physical human--robot interaction,'' \emph{Applied Mechanics Reviews}, vol.~70, no.~1, p. 010804, 2018.

\bibitem{freund2009postoptimal}
R.~M. Freund, ``Postoptimal analysis of a linear program under simultaneous changes in matrix coefficients,'' in \emph{Mathematical Programming Essays in Honor of George B. Dantzig Part I}.\hskip 1em plus 0.5em minus 0.4em\relax Springer, 2009, pp. 1--13.

\bibitem{dietrich2015overview}
A.~Dietrich, C.~Ott, and A.~Albu-Sch{\"a}ffer, ``An overview of null space projections for redundant, torque-controlled robots,'' \emph{The International Journal of Robotics Research}, vol.~34, no.~11, pp. 1385--1400, 2015.

\bibitem{chvatal1983linear}
V.~Chv{\'a}tal, \emph{Linear programming}.\hskip 1em plus 0.5em minus 0.4em\relax Macmillan, 1983.

\bibitem{bertrand_high-speed_2024}
\BIBentryALTinterwordspacing
S.~Bertrand, L.~Penco, D.~Anderson, D.~Calvert, V.~Roy, S.~{McCrory}, K.~Mohammed, S.~Sanchez, W.~Griffith, S.~Morfey, A.~Maslyczyk, A.~Mohan, C.~Castello, B.~Ma, K.~Suryavanshi, P.~Dills, J.~Pratt, V.~Ragusila, B.~Shrewsbury, and R.~Griffin, ``High-speed and impact resilient teleoperation of humanoid robots,'' in \emph{2024 {IEEE}-{RAS} 23rd International Conference on Humanoid Robots (Humanoids)}.\hskip 1em plus 0.5em minus 0.4em\relax {IEEE}, pp. 189--196.\BIBentrySTDinterwordspacing

\bibitem{orin_centroidal_2008}
\BIBentryALTinterwordspacing
D.~Orin and A.~Goswami, ``Centroidal momentum matrix of a humanoid robot: Structure and properties,'' in \emph{2008 {IEEE}/{RSJ} International Conference on Intelligent Robots and Systems}.\hskip 1em plus 0.5em minus 0.4em\relax {IEEE}, pp. 653--659.
\BIBentrySTDinterwordspacing

\bibitem{rakita2019shared}
D.~Rakita, B.~Mutlu, M.~Gleicher, and L.~M. Hiatt, ``Shared control--based bimanual robot manipulation,'' \emph{Science Robotics}, vol.~4, no.~30, p. eaaw0955, 2019.

\bibitem{corberes2025perceptive}
T.~Corbères, C.~Mastalli, W.~Merkt, J.~Shim, I.~Havoutis, M.~Fallon, N.~Mansard, T.~Flayols, S.~Vijayakumar, and S.~Tonneau, ``Perceptive locomotion through whole-body mpc and optimal region selection,'' \emph{IEEE Access}, vol.~13, pp. 69\,062--69\,080, 2025.

\end{thebibliography}
\end{document}